\documentclass[conference]{IEEEtran}
\ifCLASSINFOpdf
\else
\fi
\usepackage{graphicx}
\begin{document}
\title{Efficient Palm-Line Segmentation with U-Net Context Fusion Module\thanks{Supported by \textbf{Sun-Asterisk Inc}.}}
\author{
    \IEEEauthorblockN{Toan Pham Van, Son Trung Nguyen, Linh Bao Doan, Ngoc N. Tran}
    \IEEEauthorblockA{R\&D Lab, Sun* Inc \\
    \{pham.van.toan,nguyen.trung.son,doan.bao.linh,tran.ngo.quang.ngoc\}@sun-asterisk.com}
    
    \\
    \IEEEauthorblockN{Ta Minh Thanh}
    \IEEEauthorblockA{Le Quy Don Technical University, 236 Hoang Quoc Viet, Bac Tu Liem, Ha Noi \\
    thanhtm@mta.edu.vn}
}
\maketitle
\begin{abstract}
Many cultures around the world believe that palm reading can be used to predict the future life of a person. Palmistry uses features of the hand such as palm lines, hand shape, or fingertip position. However, the research on palm-line detection is still scarce, many of them applied traditional image processing techniques. In most real-world scenarios, images usually are not in well-conditioned, causing these methods to severely under-perform. In this paper, we propose an algorithm to extract principle palm lines from an image of a person's hand. Our method applies deep learning networks (DNNs) to improve performance. Another challenge of this problem is the lack of training data. To deal with this issue, we handcrafted a dataset from scratch. From this dataset, we compare the performance of readily available methods with ours. Furthermore, based on the UNet segmentation neural network architecture and the knowledge of attention mechanism, we propose a highly efficient architecture to detect palm-lines. We proposed the Context Fusion Module to capture the most important context feature, which aims to improve segmentation accuracy. The experimental results show that it outperforms the other methods with the highest F1 Score about 99.42\% and mIoU is 0.584 for the same dataset.
\end{abstract}

\IEEEpeerreviewmaketitle
\section{Introduction}
\subsection{Overview}
Nowadays, with great advances in computer science, image processing applications are becoming more and more popular. Palmistry is one of the interesting problems in the computer vision field. The main task of this problem is to extract the palm lines from a hand picture. The merit of this task is that it is believed the palm lines can be used to predict one's future life. Looking at these images, we can find principle lines, wrinkles, and ridges on one's palm. Usually, a hollow will have some main lines in a palm that are most notable and change little over time. Wrinkles are generally much thinner than principal lines and much more irregular. The ridge's shape is the same as the fingerprint's; hence it is difficult to distinguish them in the low-resolution image. Our task is to manually choose the best parameters for some algorithm to distinguish three line types \cite{liu2005palm}. However, this is not easy, as these parameters may work very well in some cases but not in others, resulting in a lack of generality.

As mentioned above, recent works have been focused on traditional image processing and mathematical methods. Several algorithms were applied with computer vision techniques such as noise filtering, edge detection and directional detectors \cite{kumar2017simple}\cite{smart-palmios} to accomplish this purpose. With these methods, the principle palm lines can be extracted under clear conditions and high resolution. However, the performance of previous works drops, giving low-accuracy results for images with complex background. 

To overcome these problems, deep learning approaches have been proposed for palm line extraction. An important architecture in deep learning is convolutional neural network (CNN) \cite{krizhevsky2012imagenet}. This model is useful for many tasks in computer vision such as face recognition \cite{lawrence1997face}, image classification \cite{krizhevsky2012imagenet}, image super-resolution \cite{dong2016accelerating}, semantic segmentation \cite{siam2018rtseg}, and so on. The advantage of deep learning techniques is that they are generally accurate when trained with a good dataset. That means the algorithm can cover many different cases of input images while ensuring high accuracy. Two big disadvantages when using deep learning methods are the required high-quality dataset and the low evaluation speed of the model in the production phase. To solve these two problems, in addition to building a dataset carefully, we need to build a network architecture that can take the balance of high accuracy and predictive speed of the model. We compare many network architectures to choose the optimal for both accuracy and computing speed. Our method achieves F1 score of 99.42\%, mIoU score is 0.584 and can run at 94 FPS with our mid-range consumer-tier GPU. 

\begin{figure*}[t]
    \centering
    \includegraphics[width=165mm]{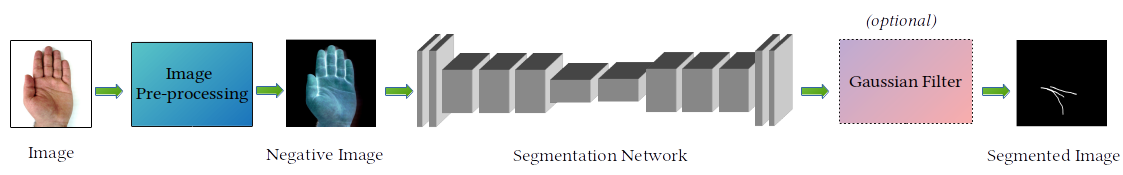}
    \caption{The architecture of our image segmentation system.}
    \label{workflow}
\end{figure*}

\subsection{Our contributions}
To sum up, our main contributions are summarized below:  
\begin{itemize}
\item(1) Proposing the use of a DNN to address the problem of palm-line detection. We use deep learning alongside image processing techniques instead of pure traditional image processing as in other previous papers.

\item(2) Providing a high-quality dataset for this problem. This dataset was carefully annotated by professionals in the field\footnote{https://link.sun-asterisk.vn/palmlinedataset}. 

\item(3) Proposing our model with the Context Fusion Module to achieve high accuracy even with complex palm-line images. With this approach, we achieve respectable performance with respect to mIoU score. 
\end{itemize}

\subsection{Roadmap}
The rest of the paper is organized as follows. Section 2 presents a brief review of related works. In Section 3, the data pre-processing method and proposed model architecture are discussed. The data preparation for the experiment and system setup is mentioned in Section 4. Experimental results and evaluation are presented in Section 5. Finally, our conclusion of the paper is in Section 6.

\section{Related works}
\subsection{Palm-line applications in real-life}

\textbf{Palm reading} (also known as palmistry) is an ancient technique originated in China. It is the analysis of a human hand to foretell the owner's future and personality. Ancient Chinese believed palm lines inhold information of humans, similar to how our ancestors found out the correlation between the movement of planets and events that happened on Earth. Be it the curve, the length, the depth or the location of the lines, every detail has its own specific meaning.

\textbf{Palm print} can be used in the field of biometric verification and recognition. Similar to fingerprinting, each person has unique palms printing. Distinct palm printing features such as geometry, lines, points, and wrinkles can be used for authentication purposes. Combined with other biometrics, the security level and privacy are increased significantly\cite{palmprint_median_filter}.

\subsection{Other approaches}

At present, numerous methods were proposed for palm detection process. However, subpar accuracy is still the main issue when extracting features of palm lines. Earlier works were affected by the limitations of pure traditional image processing techniques; and some papers suggested integrating hardware devices to improve precision, which provides specific optimal circumstances for detection. For instance, among the proposals were ROI extraction \cite{scattering-wavelet-transform}\cite{palm-print-global-features}, and 3D palmprint with structured light imaging \cite{3d-palm-print}. 


\section{Proposed method}
In this section, we present the deep learning algorithms used to solve our defined problem above. We conduct experiments with other existing deep learning models and compare its accuracy. Each architecture has advantages and disadvantages, however a majority of them suffer from the blemishes in most images with complex patterns. To overcome these difficulties, we propose a network with a custom module called Context Fusion Module (CFM) combined with the traditional U-Net architecture. This helps our network work better with input images containing complex palm printing.

\subsection{Segment Architecture}
\textbf{U-net} \cite{ronneberger2015unet} is a U-shaped convolutional neural network which was first used in the field of medical image segmentation \cite{oktay2018attention}\cite{zhou2018unet}\cite{alom2018recurrent}. It is a specific symmetric instance of the encoder-decoder network structure, with skip connections from layers in the encoder to the corresponding layers in the decoder. The encoder-decoder networks have been applied to many computer vision tasks, including object detection and semantic segmentation. These networks contain an encoder module that compresses feature maps to capture higher semantic information. And a decoder module that recovers that spatial information.



\textbf{Feature Pyramid Network (FPN)} \cite{lin2016feature} uses a standard network with multiple high spatial resolution features and adds a top-down channel with lateral connections. The top-down path begins at the deepest level of the network and is progressively upsampled while adding a converted version of the high-resolution feature from the bottom-up path. The FPN generates a pyramid, where each level has the same channel dimension.

    

\subsection{Backbone Network}

\textbf{ResNet-34} \cite{he2015deep}, or more generally, ResNet, was developed by Microsoft in 2015. It has a structure similar to VGG but with multiple stacked layers. With traditional deep learning models, extra layers are added in order to achieve better accuracy, which causes a phenomenon called vanishing/exploding gradient \cite{vanishing}\cite{philipp2017exploding} as models get deeper. ResNet with residual block are designed to solve the problem, hence providing a better outcome. ResNet-34 (34-layer) was selected among the best, based on our previous experiments.

\textbf{ResNeXt-50} \cite{xie2016aggregated} are described as a straightforward network in the task of image classification. The authors established a new hyper-parameter named cardinality, a crucial factor in addition to the model's depth and width. According to the paper, increasing cardinality shows much better results than going deeper (increasing layers) or wider (increase bottleneck width). We choose to experiment with this backbone since the authors declared that the network fares better than ResNet on both COCO detection and ImageNet-5k.

\subsection{Our network - U-Net with Context Fusion Module}

\begin{figure}[t]
    \centering
    \includegraphics[width=80mm]{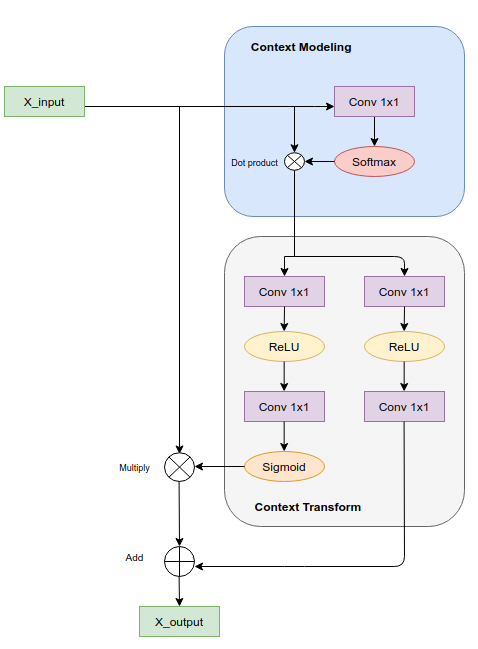}
    \caption{Context Fusion Module combining local and global context features.}
    \label{cfm}
    
\end{figure}

\begin{figure}[t]
    \centering
    \includegraphics[width=\linewidth]{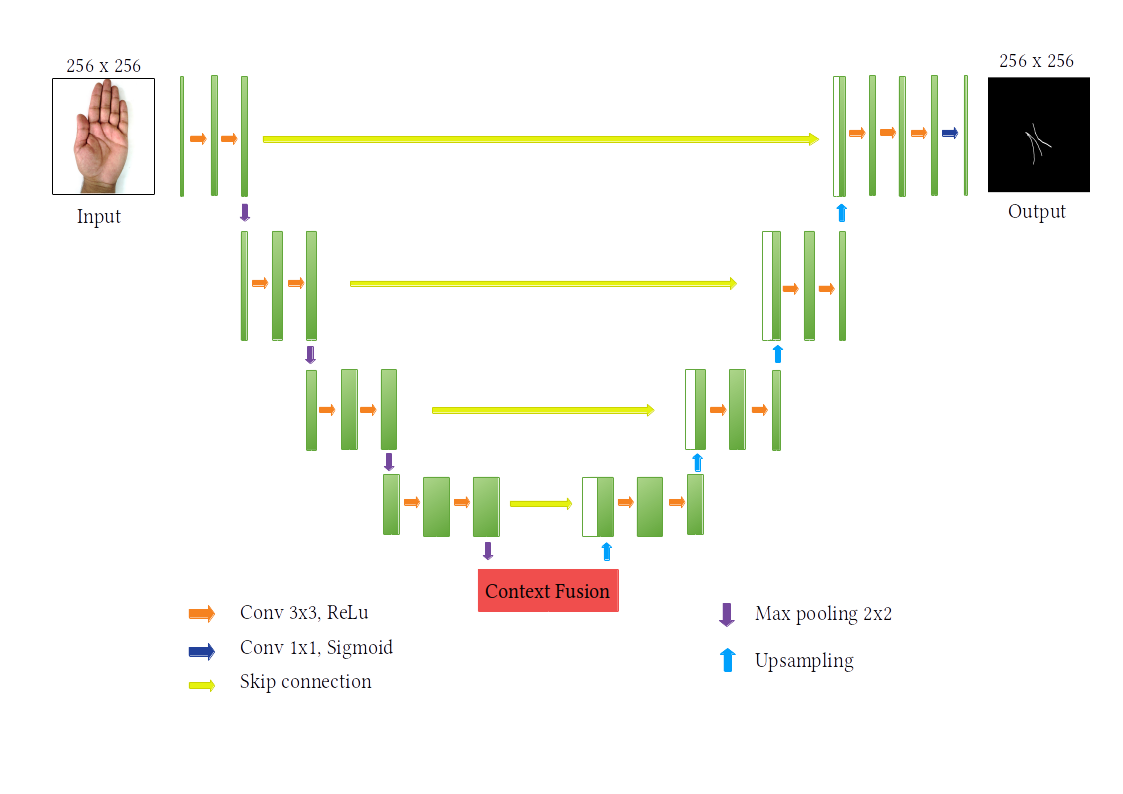}
    \caption{U-Net with Context Fusion Module}
    \label{unet_cfm}
    
\end{figure}

Attention mechanisms focus on the important regions of the local features and neglect irrelevant information of the global features. This design makes them effective in solving the long-range dependency problem. With the attention mechanism, deep learning models have become successful in many computer vision problems such as image classification \cite{wang2017residual}, image captioning \cite{you2016image}, image segmentation \cite{chen2016attention}, and so on. To minimize long-range dependencies in the palm line segmentation problem, we combine local and global features in one module called Context Fusion Module (CFM). This module is based on the attention mechanism to improve the accuracy of the overall model. It was integrated into U-Net after the encoder component as a bottleneck layer as shown in \textbf{Figure~\ref{unet_cfm}}. Our Context Fusion Module is shown in \textbf{Figure~\ref{cfm}}. It can be divided into two sub-modules. The first module called Context Modeling captures the global context features with a 1x1 convolution layer, followed by a softmax, to obtain the attention weights. The main purpose of this module is to perform attention pooling and obtain the global context features. The second module called the Context Transform module is divided into two branches. The left branch includes two 1x1 convolutions, with a ReLU activation and a sigmoid after each of them, respectively. The left branch aims to compute the importance of each channel and captures channel-wise dependencies. The right branch has the same architecture as the left branch but without the sigmoid and is independent of the left. This branch's purpose is to capture the global context feature as a piece of additional information for the fusion module. The left branch output is then used as weights to linearly combine the input before CFM yielding a local context vector, which then is fused together with the right branch through element-wise addition. The complete module is illustrated in \textbf{Figure~\ref{cfm}}. 

\section{Dataset and Training}
\subsection{Dataset}

For our task at hand, we handcrafted our own dataset.
As we mentioned in the previous part, 11K Hands dataset ($1600 \times 1200$ pixels) \cite{11khands} has 11,076 images of human hands, ranging from 18 to 75 years old. All images have same white solid background and similar distance from viewpoint.
Based on two labels ``palmar left" and ``palmar right", we gather a dataset consisting of 5243 images of palmar sides.
 We discard all but 1039 best quality images to ensure these properties of a good dataset:

\begin{itemize}

    \item \textbf{Balanced distribution.} We proceed to only select 5423 palmar side images, then filter out subpar images from high-quality ones, to remain with 512 images of the ``palmar left" label and 527 images of the ``palmar right" label.
    
    \item \textbf{Wide range variety.} The original dataset contains diverse images of skin color, gender, ages. Our custom dataset still remains the diversity of the source, achieved by carefully choose the variety in gender, skin color, hand pose, and age.
    
    \item \textbf{No incorrect labels, image noise.} There are many blurred ones, incorrectly labeled pictures or images unsuitable for instance hand with long scars, palm lines not visible, etc. We carefully observe and select the most appropriate images for the task and pass it through the upcoming annotation process.

    \item \textbf{Annotation method.} We used Supervisely \cite{supervise}, which is a professional platform for image annotation and data management. A specific tool called ``add bitmap" was utilised to draw bitmap paths along the palm lines. Also, we mainly concentrate on only visible and most meaningful lines for somatomancy purposes.
    

\end{itemize}

After the annotation process, we proceed to augmentation process with Albumentation \cite{albumentations} to diversify and enrich the dataset. The techniques we used including Horizontal-Flip, shift scale rotate, Random brightness contrast, and CLAHE. 
The final results contain 4156 images with 4156 corresponding masks that have appropriate variation in complexions, contrast, and magnitude.


\subsection{Training}
    

\subsubsection{Loss function}

Since the output of our model are probabilities denoting whether a pixel is of interest, we use binary cross-entropy as the loss function. This loss function measures the entropy difference (and equivalently, the statistical distance plus some constant) between the ground truth Bernoulli distribution and our predicted one. We sum up the pixel-level cross-entropy values to get the loss for each image as follows: 
\begin{equation}
\mathcal{L}(x',x)=-\sum_{i,j}\left(x'_{ij}\log x_{ij}+(1-x'_{ij})\log(1-x_{ij})\right)
\end{equation}
where $x'$ is the ground truth mask value (either 0 or 1), and $x$ is our predicted probability of whether a pixel is segmented as positive. Since our input size is fixed ($i=j=256$), we evaluate the class predictions for each pixel and take the average of the losses overall pixels. We also do experiments with mean squared error (MSE) as a loss candidate. However, MSE was slower and converged to a worse local minimum, which can be explained by the fact that it was a generic loss function taking no prior information about the problem into optimization.

\subsubsection{Evaluation metrics}
We opt to use 2 metrics for our model's performance:
\begin{itemize}
    \item \textbf{F1 score}: this is a natural measurement for a pixel-level classification like our models' settings. The formula for $F_1$ score is:
    \begin{equation}
    Precision = \frac{TP}{TP + FP}
    \end{equation}
    \begin{equation}
    Recall = \frac{TF}{TP+TN}
    \end{equation}
    
    \begin{equation}
    F_1=\left(\frac{2}{Recall^{-1} + Precision^{-1}}\right)
    \end{equation}
    which is the harmonic mean of the $Precision$ (true positives over predicted positives) and $Recall$ (true positives over actual positives). This gives us a better evaluation than mere accuracies in the case of imbalanced data, which happens to be our case as well, since most of the regions in the picture are not of palm lines.
    \item \textbf{IoU Score}: is a measure of accuracy for segmentation problems, defined as the ratio of the \textbf{I}ntersection region \textbf{o}ver the \textbf{U}nion region. In a way, this is the segmentation version of the recall.
    \begin{equation}
    \mathrm{IoU} = \frac{\mathrm{target} \cap \mathrm{prediction}}{\mathrm{target} \cup \mathrm{prediction}}
    \end{equation}
    
    The intersection consists of pixels in both the $target$ and $prediction$ region, while the union is the area of both have taken. The IoU score is calculated for each class separately, and then averaged over all classes to provide mean IoU (mIoU) score of semantic segmentation prediction.
    
\end{itemize}

Every image has a corresponding binary mask. The augmented dataset with a total of 4156 images was split into 3 parts: 80\% samples for training (3324 images), 10\% for validation (415 images) and 10\% samples for testing (415 images). We take experiments on our dataset in both grayscale and negative channels. As our experimental results, the negative images show the potential of giving the best results. Further inspect, we find out when an image is transformed into negative type, palm lines become brighter, so it's easier for segmentation process.

We train the vanilla U-Net, FPN and our custom network with training pair $S = \{x_i, y_i \}$ with $x_i$ is the $i$-th image and $y_i \subset \{0,1\}$ is mask (label) corresponding to $x_i$. The networks are initially set to be trained through 100 epochs with the Adam optimizer \cite{adam} and an initial learning rate of $0.0001$. Learning rate will be dropped to a fifth if the loss value does not reduce after 8 epochs. Also, if loss value still remains unchanged after 10 epochs, the training process will be automatically stopped.

With our network, the resolution of all images for training, verification and testing will be resized to $256 \times 256$ because model will train faster with smaller images and using less memory and computational power, which fit
for our limited resources. Batch size for all processes is fixed to 64. The loss function is the binary cross-entropy as we have mentioned above. We train the network for slightly more than two hours with early stopping at epoch number 36.

\subsection{System configuration}
Our experiments are conducted on a computer with Intel Core i5-7500 CPU @3.4GHz, 32GB of RAM, GPU GeForce GTX 1080 Ti, and 1TB SSD hard disk. The models are implemented with the TensorFlow framework \cite{abadi2016tensorflow} and Keras API \cite{chollet2015keras}.
\section{Result comparisons}

\subsection{Results}


\begin{figure}[t]
    \centering
    \includegraphics[width=\linewidth]{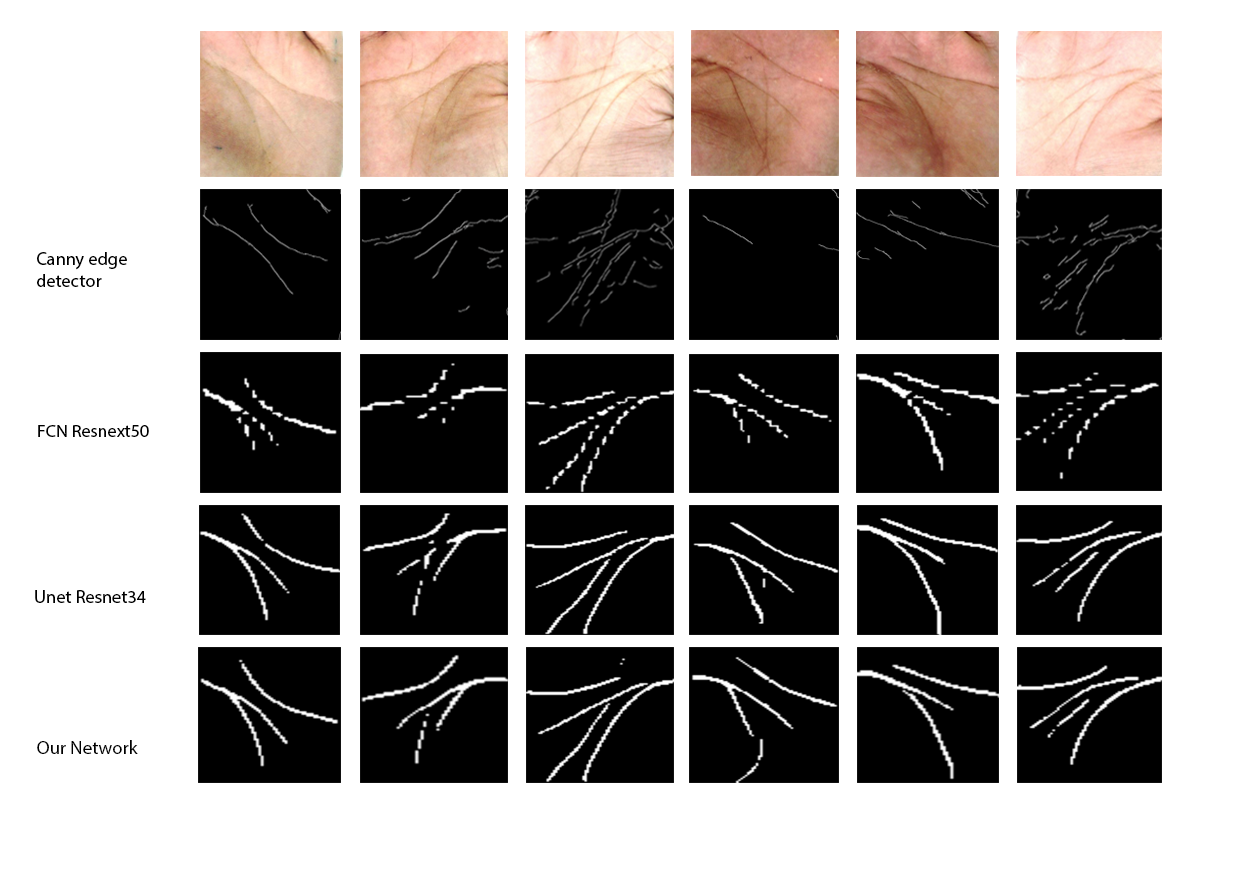}
    \caption{The output images of some models applied to this problem. Our network (Unet-CFM) achieve a better result with complex palm line input.}
    \label{compare_masks}
\end{figure}

In \cite{smart-palmios}, the authors proposed a system that can handle hand images after various preprocessing techniques like desaturation, threshold, dilation and palm extraction by locating special points, applying interpolation, and determining the Region of Interest. With the extracted palm hand, they continued to use Canny edge detector to get the palm lines. Canny edge detector, which was developed by John F. Canny in 1986 \cite{canny}, is a multi-stage algorithm, designed for finding the edge regions. We have successfully implemented the proposed algorithm in \cite{smart-palmios} for comparison along with U-Net or FPN (with ResNet-34 and ResNeXt-50 as the backbone), and our network. \textbf{Figure \ref{compare_masks}} gives more detail of the experimental results.

    

\textbf{Table \ref{mean_iou}} shows the numerical result comparisons. For mIoU, our model surpasses U-Net with 34-layer baseline ResNet backbone by 0.045 and ResNeXt-50 backbone by 0.049. Against FPN, U-Net-CF shows superior improvement with a 0.228 difference on ResNet-34 and 0.193 on ResNeXt-50. F1 Score also reveals 0.53\% and 0.41\% increment compare to U-Net; 3.8\% and 3.41\% gain compare to FPN. Our proposed model has fewer parameters (only 10,270,115) than other deep networks such as ResNet-34 and ResNeXt-50. Nevertheless, as shown in the table, our network still outperforms other approaches. Therefore, applying CFM can intuitively be considered as a promising method. It provides a new approach for palmprinting algorithm that in the future can be investigated for localization performance improvements.

\begin{table}[h!]
\caption{ Quantitative comparison between U-Net, FPN and U-Net-CF }
\centering
 \begin{tabular}{| c | c | c | c | c |} 
 \hline
  \textbf{Method} & \textbf{Backbone} & \textbf{Params} &\textbf{F1 Score} &\textbf{mIoU}  \\ [0.5ex] 
 \hline
  Unet &ResNet-34  &24,456,299   &98.89\%  &0.539  \\
       &ResNeXt-50 &32,063,339  &99.01\%  &0.535 \\
  \hline
  FPN  &ResNet-34 &25,696,459  &95.62\%  &0.356 \\
       &ResNeXt-50 &28,179,403  &96.01\%  &0.391 \\
  \hline
  Unet-CF & Unet \cite{ronneberger2015unet} &\textbf{10,270,115}  &\textbf{99.442\%}  &\textbf{0.612584} \\
 \hline
 \end{tabular}
\label{mean_iou}
\end{table}

\subsection{Gaussian Filter}

 Gaussian Filter \cite{gauss} are usually used to generate blur image. Our main purpose of using this filter is to reduce image noise or any excessive detail. Gaussian filter's basic idea is that a pixel's location affects its density in the image. For instance, pixels located in the middle would have the biggest weight. The weight of its neighbors decreases as the spatial distance between them and the center pixel increases:
 $$G_{0}(x, y) = A e^{ \frac{ -(x - \mu_{x})^{2} }{ 2\sigma^{2}_{x} } + \frac{ -(y - \mu_{y})^{2} }{ 2\sigma^{2}_{y} } } $$
 with $\mu$ being the mean (peak) and $\sigma $ the variance of $x$ and $y$. Parameter $\mu$ controls the amount of change the Gaussian filter act upon the image. The size of kernel should be chosen wide enough, which we selected to be $3\times3$ for this process.
 
 \begin{figure*}[t]
    \centering
    \includegraphics[width=122mm]{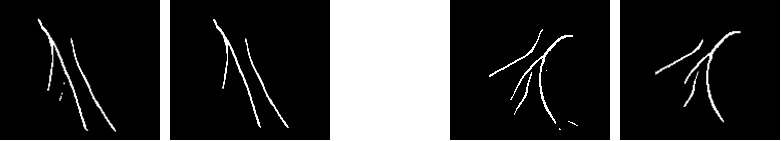}
    \caption{Results with (right) and without Gaussian blur (left).}
    \label{blur}
\end{figure*}

\textbf{Figure \ref{blur}} shows the result with and without the Gaussian blur post-processing step. As expected, this optional step reduces disconnected random pixels being classified as of interest, but at the same time smoothens the edges of our segmentation output. As a result, the mIoU actually decreases, albeit by a very small margin. One may consider opting to include this step if you want a nicer-looking mask, for example if the next task in your pipeline requires it.

\section{Future works}
Our research opens a plethora of possibilities to be considered. For example, one may be interested in exploring the idea of applying the Context Fusion Module on each of the skip connection in every tier of the U-Net architecture. Or, they can experiment with directing the encoding output of each U-Net tier into the CFM for it to have a more comprehensive interpretation of the information, which then could be distributed back to the respective decoding parts. Another direction to be considered is whether replacing U-Net with Feature Pyramid Network (FPN)-like structure would be a good idea: CFM would work great with pyramid-pooling scheme; however traditionally FPN models are used for object detection tasks instead of segmentation. Along with experimenting with other models, we can also improve our dataset with either better processing to handle the variations of complicated background images, or simply increases the amount of data for our model to learn.

\section{Conclusion}
In this paper, we applied deep learning techniques to build neural networks to solve the palm lines segmentation problem. The final mIoU of our model is 0.584 and F1 score is 99.42\% on our dataset. This dataset was collected manually and will be distributed publicly for scientific purposes. The experimental results show that the proposed method has tremendous advantages over the traditional image processing in the palm-line image segmentation tasks. Future works of the present study would be working on a more robust method to handle the variations of complicated background images; also further investigations can be done using other functionalities of the CFM.

\section*{Acknowledgment}
This work is partially supported by \textbf{\textit{Sun-Asterisk Inc}}. We would like to thank our colleagues at \textbf{\textit{Sun-Asterisk Inc}} for their advice and expertise. Without their support, this experiment would not have been accomplished.
\bibliographystyle{IEEEtran}
\bibliography{references}

\end{document}